\documentclass[education,article,submit,oneauthor]{mdpi}

\let\linenumbers\relax

\firstpage{1}
\pubvolume{1}
\issuenum{1}
\articlenumber{0}
\pubyear{2026}
\copyrightyear{2026}
\datereceived{}
\daterevised{}
\dateaccepted{}
\datepublished{}

\Title{A Risk-Adaptive and Evidence-Constrained Framework for Generative AI Feedback in Programming Education}

\Author{Shihao Wang $^{1,*}$}
\AuthorNames{Shihao Wang}

\address{$^{1}$ \quad Department of Education, Practice and Society, UCL Institute of Education, University College London, London, United Kingdom; shihao.wang.24@ucl.ac.uk}

\corres{Correspondence: shihao.wang.24@ucl.ac.uk}

\abstract{Generative artificial intelligence can turn learning analytics into personalized support, but feedback systems must decide when to intervene, which evidence to use, and how much assistance to provide. We developed a risk-adaptive, evidence-constrained framework for introductory programming using 2993 failed-submission states from 215 students. Student-disjoint models predicted persistent failure and related outcomes; four matched feedback conditions were generated for 136 cases; and calibrated risk informed capacity-limited intervention policies. The validation-selected logistic regression model achieved a test precision--recall area under the curve of 0.550 and a receiver operating characteristic area under the curve of 0.681. Broader student histories improved prediction of unmodified resubmission. After standardized repair and evidence gating, 519 of 544 newly generated messages contained all required components. A fixed-threshold sequential policy selected 17.8\% of eligible test states and captured 25.2\% of observed persistent failures. These findings support an evidence-gated progressive assistance strategy: calibrated risk guides intervention timing, recorded evidence constrains feedback content, and assistance progresses from self-checks to localized hints when warranted. The framework connects prediction, decision-making, and grounded generation while keeping their evaluation outcomes distinct.}


\begin{document}

\section{Introduction}

Feedback connects current performance with the actions needed for improvement, but its effectiveness depends on content, timing, and design \citep{wisniewski2020feedback}. These decisions are especially important in introductory programming, where students repeatedly write code, run tests, interpret failures, and revise solutions. Each failed submission creates an opportunity for support, yet the appropriate response depends on the learner's recent history and available evidence. Effective feedback must therefore determine whom to support, when to intervene, what evidence to use, and how much guidance to provide.

Learning analytics can inform these decisions through behavioral and performance traces. Higher-education research shows that such data can support risk identification and targeted intervention, while dashboards and open learner models can link indicators to reflection and learner action \citep{ifenthaler2020learning,matcha2020dashboards,hooshyar2020open}. Generative AI extends this capability by converting contextual evidence into natural-language scaffolds \citep{kasneci2023chatgpt,molenaar2022hybrid,li2025adaptive}. Instructional guardrails are important because support should promote independent reasoning alongside immediate task progress \citep{bastani2025guardrails}.

Programming submissions provide precise evidence from code, tests, and revisions. Recent studies have generated code explanations, novice-oriented error messages, and hints with varying levels of specificity \citep{sarsa2022generation,leinonen2023errors,xiao2024hints,lohr2025types}. Their evaluations demonstrate technical feasibility and identify correctness, completeness, comprehensibility, and repair accuracy as key quality dimensions \citep{koutcheme2025evaluating}. Two design questions remain closely connected: when feedback should be triggered within an attempt sequence and how learner history should regulate its content and specificity.

These questions require prediction and feedback generation to be treated as parts of a single educational process. A risk score is useful only when it informs an appropriate decision under realistic limits on instructor attention or automated support. Likewise, a well-written message is educationally credible only when its claims are grounded in the learner record available at that moment and its specificity matches the required level of assistance. Integrating timing, grounding, and assistance intensity therefore provides a stronger basis for personalized feedback than optimizing prediction accuracy or message fluency separately.

This integration also clarifies how the three intended learning outcomes can be examined. Knowledge mastery and longer-term performance require predictions linked to later task evidence, whereas self-regulated learning requires indicators of monitoring, revision, and strategy use rather than a single correctness score. A unified framework can preserve these distinctions while using the same temporally ordered evidence to determine when feedback is likely to be useful.

Ordered programming traces provide the temporal evidence needed to address these questions \citep{price2020progsnap2}. Sequence models can represent evolving learner histories, while interpretable models provide transparent benchmarks. Credible comparison requires student-disjoint evaluation and predictors restricted to information available at the intervention point \citep{kapoor2023leakage}. Prediction must then be connected to generation: calibrated risk can guide intervention timing and assistance level, while bounded task and code evidence can ground diagnostic statements and enable source verification \citep{huang2025hallucination}.

The present study develops a risk-adaptive and evidence-constrained framework for this purpose. It analyzes 2993 failed-submission states from 215 students and predicts persistent failure, unmodified resubmission, near-term related-task performance, and 7--28-day related-task performance. Logistic regression and gradient boosting are compared with multilayer perceptron, gated recurrent unit, Transformer, fusion, and multitask models using student-level partitions. A paired generation experiment represents 136 cases under four contextual conditions while holding the language model constant and varying current-task evidence, recent history, module context, calibrated risk, and assistance intensity. Fixed-capacity prioritization and sequential threshold replay assess intervention timing, while structural and source checks assess generated feedback.

The study is guided by four research questions:

\begin{enumerate}[label=\textbf{RQ\arabic*:}, leftmargin=*]
    \item How accurately can current-submission evidence and prior learning traces predict near-term persistent failure and the defined operational outcomes under student-disjoint evaluation?
    \item How do interpretable tabular baselines and deep sequential models compare in predictive performance, calibration, and the contribution of different evidence groups?
    \item To what extent can four progressively contextualized feedback conditions generate structurally complete, actionable, and source-traceable messages?
    \item Under fixed-capacity and sequential decision settings, how effectively can calibrated risk estimates identify timely opportunities for progressively adjusted assistance?
\end{enumerate}

This study contributes an auditable link between learner-state prediction and feedback generation. It examines the trade-off between predictive value and model complexity, introduces a paired design for evaluating contextual evidence and assistance intensity, and frames intervention timing as a capacity-aware educational decision. Source matching strengthens diagnostic traceability, while the progressive assistance policy translates calibrated risk into a structured sequence of self-checks, targeted hints, and more explicit guidance. Together, these elements provide a reproducible basis for personalized, explainable, and timely learning support and for evaluating knowledge mastery, self-regulated learning, and longer-term performance.

\section{Literature Review}

\subsection{Feedback as a Personalized and Timely Learning Process}

Feedback supports learning when students can connect it to a current goal and use it in subsequent action. Its effectiveness varies with message content, task characteristics, learner needs, and implementation context \citep{wisniewski2020feedback,morris2021formative}. Personalization therefore requires both evidence about the learner and a pedagogical rationale for selecting a response. Reviews of digital learning environments show that most systems adapt to current knowledge or observed behavior, while goals, affect, and progress over time are incorporated less consistently \citep{maier2022personalized}. In programming, meaningful adaptation may vary the location, conceptual depth, or specificity of a hint according to the current attempt and recent revisions.

Timing should likewise reflect the learning situation. Early support can prevent repeated unproductive attempts, whereas premature intervention can disrupt productive struggle. An effective policy should respond to evidence of continuing difficulty rather than impose a uniform delay. Assistance intensity must be regulated for the same reason. Programming studies show that learners respond differently to broad prompts, localized hints, and explicit suggestions \citep{xiao2024hints,lohr2025types}. Evidence from mathematics further suggests that unrestricted generative assistance can improve supported practice while reducing later unassisted performance, whereas instructional guardrails can mitigate this effect \citep{bastani2025guardrails}. Adaptive feedback should therefore coordinate intervention timing with progressively adjusted support.

Feedback content should also preserve an active role for the learner. A broad self-check may be sufficient when progress is evident, while repeated failure may justify a concept reminder or localized hint. More explicit guidance should be reserved for sustained difficulty. This progression links personalization with formative purpose: assistance changes with observed need, while each message directs attention toward interpretation, revision, and verification. It also offers a practical basis for evaluating whether support remains useful without revealing more of the solution than the situation requires.

\subsection{Learning Analytics, Learner Models, and Self-Regulated Learning}

Learning analytics organizes interaction traces into indicators of performance, effort, and change over time. Higher-education research shows that these indicators can identify risk, visualize progress, and guide attention toward useful next actions \citep{ifenthaler2020learning,banihashem2022analytics}. Dashboards and open learner models can make such evidence visible and support reflection, particularly when indicators are linked to strategies learners can enact \citep{matcha2020dashboards,hooshyar2020open,paulsen2024dashboards}. Generative AI extends this approach by translating selected indicators into contextualized explanations and scaffolds \citep{li2025adaptive}.

Digital traces describe observable actions rather than internal states. Submission timing, test outcomes, revisions, and inactivity can indicate monitoring or persistence, but their interpretation depends on surrounding events. Trace-based research on self-regulated learning is strongest when indicators are linked to a theoretical construct and a defined temporal window \citep{du2023traces}. Accordingly, an unchanged resubmission can represent a specific revision behavior, while broader claims about self-regulation require additional evidence. Similarly, later task performance can operationalize near-term transfer or longer-term success without being equated with complete knowledge mastery.

This distinction also shapes explainability. Learner-state representations should expose the observations supporting a decision rather than only a risk score. Failed checks, revision counts, elapsed time, and prior attempts can justify intervention urgency or assistance level, while claims about specific misconceptions should be grounded in task or code evidence. This approach follows open learner model research by making analytics actionable and inspectable \citep{hooshyar2020open}.

Learning analytics also distinguishes prediction from pedagogical interpretation. A model may identify a high-risk state from weak scores, repeated failures, and limited code change, but the estimate alone does not explain why the learner is struggling. The intervention layer must translate that estimate into a bounded decision: whether to provide support, which recorded evidence to use, and what level of assistance is appropriate. This separation improves auditability and prevents statistical associations from being presented as diagnoses. It also allows different outcome models to inform immediate recovery, revision behavior, related-task performance, and longer-term performance.

\subsection{Sequential Modeling of Learning Processes}

Programming-process data preserve the order of attempts, edits, tests, and task transitions, enabling analysis of how a learner reached a given state \citep{price2020progsnap2}. This temporal order matters because identical failed outputs may emerge from different histories and therefore warrant different forms of support. Knowledge tracing research has accordingly progressed from probabilistic approaches toward recurrent, memory-based, graph-based, and attention-based models \citep{abdelrahman2023knowledge}. Gated recurrent units summarize variable-length histories through learned gating mechanisms \citep{cho2014gru}, whereas Transformer encoders use self-attention to model dependencies across sequence positions \citep{vaswani2017attention}.

The present task extends conventional knowledge tracing by predicting whether an observed programming failure persists across the next two attempts. Recurrent and attention-based models test the value of ordered history, while logistic regression and gradient boosting provide structured-data baselines. Model selection should consider calibration and transparency alongside discrimination because estimated probabilities determine intervention priority. Student-disjoint partitions, training-only preprocessing, and prefix-restricted inputs are therefore essential, as random event splits or future-derived features can substantially inflate performance \citep{kapoor2023leakage}. Validation-fitted probability calibration further aligns predicted risk with capacity-aware decisions \citep{guo2017calibration}.

These model families provide complementary evidence. Tabular baselines show how much can be learned from the current state and summarized history with relatively direct interpretation. GRUs test whether gated recurrence captures meaningful progression across attempts, while Transformers test whether attention across positions improves representations of longer dependencies. Their comparison is informative only when preprocessing, data partitions, outcome definitions, and selection rules remain fixed. Under limited instructional capacity, calibrated probabilities are especially useful because validation-selected thresholds can be translated into defined intervention volumes rather than treated as abstract model scores.

\subsection{Generative AI Feedback in Programming Education}

Large language models have extended programming support beyond fixed templates to natural-language explanations and hints. Prior work has demonstrated automated exercise and explanation generation, novice-oriented error messages, multi-level hints, and specified feedback types \citep{sarsa2022generation,leinonen2023errors,xiao2024hints,lohr2025types}. These studies establish technical feasibility, but feedback quality still varies across tasks, prompts, and evaluation criteria.

Evaluation has therefore moved beyond fluency. Recent studies examine correctness, completeness, comprehensibility, and repair accuracy, while also showing that language models remain imperfect judges of generated feedback \citep{koutcheme2025evaluating}. Research on writing feedback likewise reports benefits for revision and selected quality dimensions \citep{meyer2024feedback,steiss2024quality}. Across higher education, systematic evidence highlights the potential of generative feedback alongside the importance of accuracy, transparency, learner agency, and instructor oversight \citep{lee2024genai}. Message quality and learning effectiveness should therefore be evaluated as distinct outcomes.

Grounding is especially important when feedback refers to code, tests, or learning history. A plausible response may still contain an unsupported diagnosis, so readability alone cannot prevent hallucination \citep{huang2025hallucination}. An evidence-constrained generator should operate on a bounded learner record, distinguish observations from inferences, recommend a feasible next action, and preserve traceability to quoted sources. Explainability should also match stakeholder needs: feature attribution can support researcher audit \citep{lundberg2017shap}, whereas learners need concise explanations linked to their next action and instructors may require fuller records of risk, evidence, and uncertainty \citep{khosravi2022explainable}.

Evaluation design should isolate the effect of contextual information from differences among cases or models. A within-case comparison can hold the task and generator constant while adding recent history, module context, calibrated risk, or an assistance instruction. Structural checks can assess whether responses contain an observation, an actionable step, and a self-check, while source verification can test whether quoted code appears in the supplied record. Human review can further assess correctness, relevance, and pedagogical appropriateness. Together, paired generation and explicit evidence checks clarify how additional context changes feedback and which messages warrant further review.

\subsection{Synthesis and Research Gap}

The literature establishes the core components of adaptive feedback but typically evaluates them in isolation. Feedback research examines timing and assistance; learning analytics provides behavioral evidence; sequential models estimate evolving risk; and generative models express support in natural language. A deployable system must connect these functions by identifying intervention opportunities, selecting time-valid evidence, regulating feedback specificity, and preserving an auditable record of each message.

Emerging work has begun to integrate real-time analytics with generative scaffolding \citep{li2025adaptive}, yet calibration, intervention capacity, grounding, and progressive assistance are rarely examined together. The present framework connects observation, prediction, decision, generation, and evaluation through student-disjoint prediction, calibrated capacity-aware policies, bounded generation contexts, and source checks. This integration supports personalized, explainable, and timely feedback while establishing a clear basis for subsequent evaluation of knowledge mastery, self-regulated learning, transfer, and retention.

This integration also makes each claim testable at the appropriate stage. Predictive validity concerns held-out outcomes and calibration; policy value concerns the precision and coverage of selected intervention opportunities; and feedback quality concerns structure, actionability, and traceability. Keeping these endpoints connected but distinct enables a more rigorous assessment of the overall feedback strategy.

\section{Methodology}

\subsection{Research Design and Scope}

This study used a retrospective computational design combining learner-state prediction, paired feedback generation, and historical-log policy evaluation. The unit of analysis was a failed programming state at which feedback could be considered. The framework comprised five linked stages: (1) construct a time-valid representation from the current submission and prior learning traces; (2) estimate the risk of a defined future outcome; (3) map calibrated risk to an intervention decision under fixed feedback capacity; (4) generate feedback from bounded source evidence; and (5) evaluate prediction, calibration, policy capture, and observable message properties. This structure aligns each research claim with a corresponding evaluation endpoint.

The analyses addressed two complementary components. The predictive component examined whether current-state and sequential features could identify students likely to remain unsuccessful across their next two attempts and predict related behavioral and performance outcomes from the same traces. The generation component examined whether progressively richer, source-bounded context changed feedback structure and traceability while holding the language model and decoding procedure constant. Intervention timing was evaluated by replaying decision rules over recorded trajectories and comparing the opportunities selected by alternative policies.

\begin{figure}[htbp]
\centering
\includegraphics[width=\textwidth]{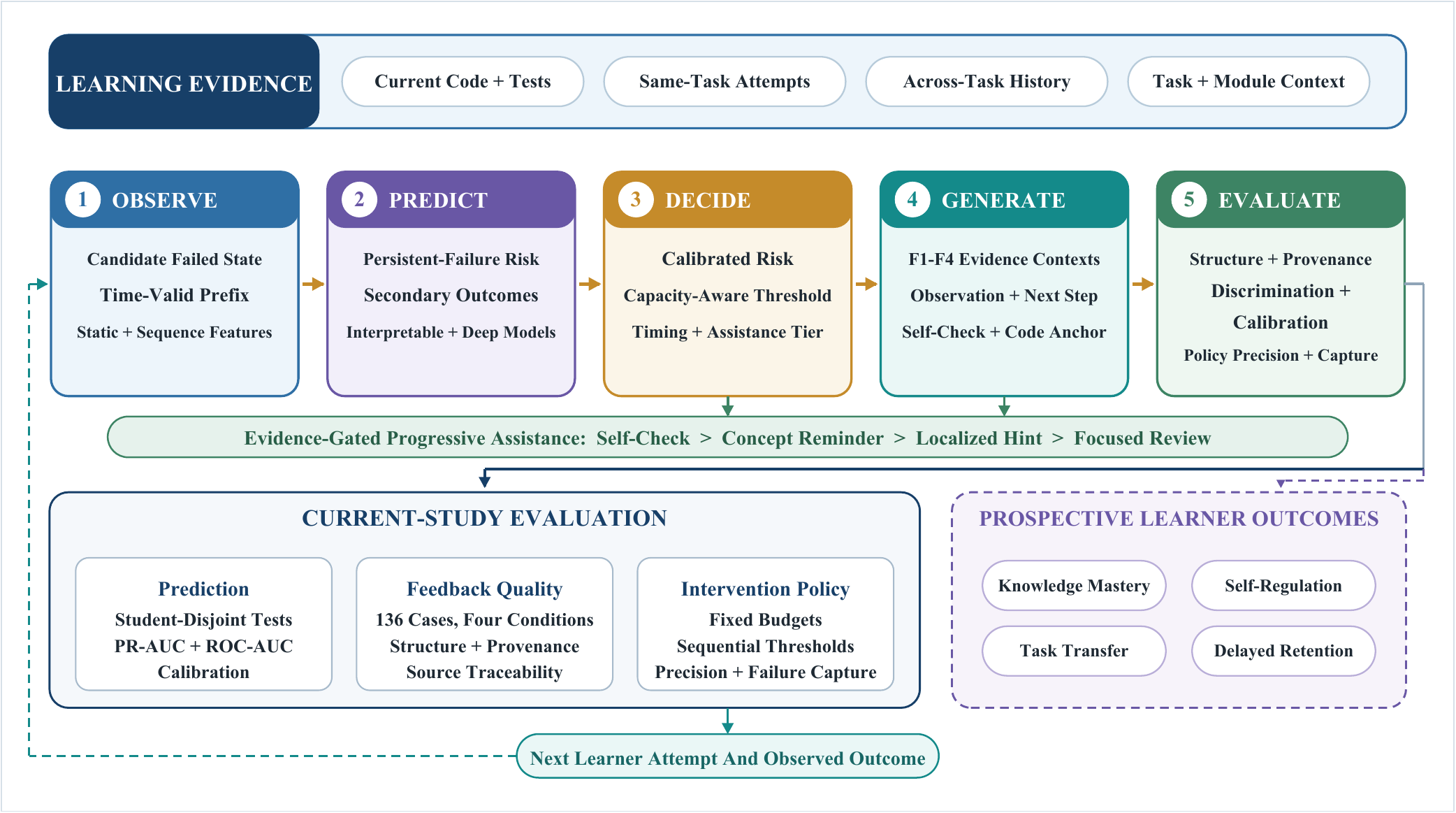}
\caption{Overall research framework.}
\label{fig:framework}
\end{figure}

Figure~\ref{fig:framework} summarizes the complete framework. Learning evidence progresses through observation, prediction, decision, generation, and evaluation, while the evidence-gated progressive assistance policy links calibrated risk to feedback specificity. Each subsequent attempt contributes new evidence for the next decision, creating a sequential cycle of adaptive support.

\subsection{Dataset, Participants, and Analytical Cohort}

The study used the public ProgFeed dataset, a de-identified record of an introductory programming course delivered in Fall 2025 \citep{umass2025progfeed}. The repository includes programming submissions, autograder results, problem statements, recorded feedback conditions, and entry and exit surveys. Its documentation states that only students who consented to research use were included and that direct identifiers were removed. The data are released under a CC BY 4.0 license. A fixed repository snapshot was used throughout the analysis.

The consolidated source contained 17,385 graded function--test records from 215 students, representing 6693 submissions and 16,365 function-level submission states. Records sharing student, laboratory, source file, function, and timestamp were aggregated into a single state after confirming one code version and nonduplicated test identifiers. Test scores and maximum scores were summed within each state, and any failed functional check classified the state as unsuccessful. Student code was parsed as text for structural features but was never executed.

A candidate decision state required an unsuccessful attempt with nonempty code, a preceding attempt on the same task, and inclusion in the de-identified research dataset. These criteria produced 2993 candidate states. Students were assigned once to training, validation, or test partitions using a reproducible 60\%/20\%/20\% split, with all records from each student retained in the same partition. Table~\ref{tab:cohort} summarizes the cohort and primary-outcome coverage. The 238 endpoint-censored states were retained in prediction and deployment-oriented records, while supervised performance estimates used the 2755 states with observed outcomes.

\begin{table}[htbp]
\caption{Student-disjoint partitions and primary-outcome availability.\label{tab:cohort}}
\centering
\resizebox{\textwidth}{!}{%
\begin{tabular}{lrrrrr}
\toprule
Partition & Students & Candidate states & Known outcome & Unknown outcome & Persistent failure \\
\midrule
Training & 129 & 1897 & 1756 & 141 & 847 \\
Validation & 43 & 499 & 444 & 55 & 183 \\
Test & 43 & 597 & 555 & 42 & 222 \\
\midrule
Total & 215 & 2993 & 2755 & 238 & 1252 \\
\bottomrule
\end{tabular}}
\end{table}

The recorded A--D feedback subset comprised 136 matched cases and 544 messages, each with a human-reviewed overall rating on a 0--100 scale. A separate generation experiment used the same cases to produce 544 F1--F4 messages, evaluated through structural and source checks.

\subsection{Outcome Construction and Measurement Boundaries}

The primary outcome, \emph{next-two-attempt persistent failure}, was defined from the next two chronological states for the same student and task. The outcome was coded 0 if either attempt passed and 1 if both observed attempts failed. When fewer than two subsequent attempts were available and no pass was observed, the outcome remained missing. Future states were used only to construct outcomes and were never supplied as predictors or generation context.

Three secondary outcomes were derived from later observed records. An \emph{unmodified retry} indicated that the next same-task submission had the same canonical abstract syntax tree (AST) as the current code. If either version could not be parsed, code normalized for comments and whitespace was compared instead. This variable captures an observable revision behavior related to monitoring and strategy change. For the two task-performance outcomes, tasks were assigned before model fitting to seven broad content groups: basic input/output and arithmetic, conditional logic, iteration, collections, file input/output, classes and state, and recursion. \emph{Near-term related-task performance} recorded functional success on the first future distinct task in the same broad group within seven days. \emph{Longitudinal related-task performance} used the corresponding first observation after 7 days and within 28 days. When no qualifying future task was observed, the label was coded as missing.

Table~\ref{tab:outcomes} summarizes the operational definitions and coverage. The two related-task outcomes capture transfer-like and longitudinal performance under ordinary course conditions. Their explicit content and temporal definitions support theory-aware interpretation of trace-derived indicators \citep{du2023traces} and provide observable targets for the computational evaluation.

\begin{table}[htbp]
\caption{Operational outcomes used in the predictive analyses.\label{tab:outcomes}}
\centering
\begin{tabular}{p{3.1cm}p{6.9cm}rr}
\toprule
Outcome & Operational definition & Cases & Students$^{a}$ \\
\midrule
Persistent failure & Both of the next two same-task attempts failed; a pass in either attempt was coded as nonpersistent. & 2755 & 198 \\
Unmodified retry & Next same-task code had an identical canonical AST, with normalized-text fallback for unparseable code. & 2847 & 198 \\
Near-term related-task performance & Functional success on the first distinct task in the same broad content group within 7 days. & 950 & 147 \\
Longitudinal related-task performance & Functional success on the first distinct same-group task after 7 days and within 28 days. & 307 & 105 \\
\bottomrule
\end{tabular}
\begin{flushleft}
\footnotesize{$^{a}$ Student counts are summed across disjoint partitions and may include the same student in more than one outcome row.}
\end{flushleft}
\end{table}

\subsection{Time-Valid Feature Construction}

Static predictors described the learner state at the decision time. They included attempt count; current and previous score ratios; score change; numbers of failed and total checks; failures across the three most recent attempts; consecutive failures; elapsed time since the previous attempt; code length; edit fraction; and a syntax-parseability indicator. Ten additional counts captured AST structure: total nodes, calls, conditional statements, \texttt{for} and \texttt{while} loops, returns, binary operations, comparisons, function definitions, and literals. Laboratory, source file, and function identifiers were represented categorically. Student identifiers were used only for partitioning and clustered evaluation and never as model inputs.

Continuous missing values were imputed with training-set medians and paired with explicit missingness indicators. Positively skewed count and time variables were transformed using \(\log(1+x)\), after which continuous features were standardized with means and standard deviations estimated from known-outcome training cases. Categorical vocabularies were fitted on the training partition, with unseen validation or test categories mapped to an unknown level before one-hot encoding. AST features were derived through parsing only; parsing failures remained missing and were represented by indicators. No student or autograder program was executed during preprocessing.

The primary sequence representation contained up to 20 states from the same student, laboratory, source file, and function, ending at the decision state. Each timestep comprised eight values---score ratio, failed-check count, total-check count, failure status, attempt index, elapsed time, score change, and consecutive failures---together with eight missingness indicators. Transformations and normalization were fitted only on sequence states within training prefixes. Shorter sequences were padded within batches, with true lengths supplied to the recurrent model or used to construct a Transformer padding mask.

A follow-up across-task sequence examined whether broader student history improved prediction beyond same-task prefixes. It included the latest 20 completed states for the same student across tasks, restricted to timestamps no later than the current state. Each 16-dimensional timestep combined eight normalized numeric variables, seven one-hot content-group indicators, and a score-ratio missingness indicator. These analyses are reported as validation-informed extensions of the primary representation.

\subsection{Predictive Models and Training Procedure}

Eight model configurations were compared for the primary outcome. Logistic regression and histogram gradient boosting served as validation-tuned tabular baselines. The deep-learning comparison included a multilayer perceptron (MLP), a sequence-only gated recurrent unit (GRU), a static--sequence fusion GRU, the same fusion model without AST predictors, a fusion Transformer, and a multitask fusion GRU. GRUs use learned gates to retain relevant sequential information \citep{cho2014gru}, whereas Transformers model cross-position relations through self-attention \citep{vaswani2017attention}. Their inclusion also reflects the broader knowledge-tracing literature, although the present target is observed future failure rather than latent mastery \citep{abdelrahman2023knowledge}.

For the MLP and fusion models, the static vector was projected to 32 units with rectified linear activation and dropout of 0.20. The GRU used a 16-dimensional timestep input and a hidden size of 32. In the Transformer, the timestep vector was projected to 32 dimensions and combined with a learned positional embedding for 20 positions. A single encoder layer used four attention heads, a feed-forward dimension of 64, and dropout of 0.20. Fusion models concatenated the 32-dimensional static and sequential representations, followed by a prediction head with a 32-unit hidden layer, rectified activation, dropout of 0.20, and a sigmoid output. The MLP used only the static representation, while the sequence-only GRU omitted the static branch.

Deep models were trained with AdamW \citep{loshchilov2019adamw} and validation-based early stopping on precision--recall area under the curve (PR-AUC). Training was repeated with five random seeds, and predicted probabilities were averaged across runs. The multitask model jointly predicted all eligible outcomes, assigned greater weight to the primary outcome, and masked missing labels from the corresponding secondary losses.

Hyperparameters, early stopping, and model selection used only the validation partition. The test set was reserved for final performance estimation and was not used for model selection. Logistic regression, gradient boosting, and separate fusion-GRU models were also fitted for each secondary outcome using all cases with an observed label for that outcome, including states with missing primary outcomes. For the follow-up across-task sequence, a multitask fusion GRU and outcome-specific GRUs were trained using the same optimization procedure. Summary-feature logistic and gradient-boosting baselines used the same across-task history source to separate gains from sequence modeling from gains due solely to broader history.

For probability calibration, Platt scaling was fitted to validation predictions from the selected deep ensemble by applying logistic regression to the logit-transformed ensemble probability \citep{guo2017calibration}. The calibrator and all decision thresholds were fixed before test-set evaluation. The across-task multitask GRU used a separate validation-fitted calibrator for the risk-adaptive strategy.

\subsection{Predictive Evaluation and Interpretation}

PR-AUC, computed as noninterpolated average precision, was the primary discrimination metric because persistent failure was unevenly distributed. ROC-AUC, Brier score, logarithmic loss, F1 score at a 0.50 probability threshold, and balanced accuracy were also reported. Calibration was assessed using quantile-binned reliability curves. To account for repeated states within learners, 95\% confidence intervals were estimated from 1000 student-level bootstrap samples. Paired bootstrap differences used the same resampled students for each model comparison. Student-disjoint evaluation and time-restricted predictors reduced the risk of within-student and temporal leakage \citep{kapoor2023leakage}.

Interpretability analyses were separated from learner-facing explanations. Logistic coefficients represented conditional associations in the standardized feature space. For gradient boosting, features were grouped into score and failure, timing and attempts, code edits, AST structure, and task identifiers; each group was permuted 30 times on the test set, and the resulting change in PR-AUC was recorded. Attention weights were not treated as explanations. These analyses supported model audit, whereas learner-facing feedback was grounded in observable task or code evidence linked to the next action.

\subsection{Paired Feedback Generation and Progressive Assistance}

The generation experiment used the same 136 cases across four within-case conditions. F1 supplied the task statement, current submitted code, and recorded failed checks. F2 added the two most recent states and compact history indicators, including changes in failure counts and whether the code had changed. F3 added the complete current module to support checks of dependencies and code locations. F4 combined F2 and F3 with calibrated persistent-failure risk from the across-task multitask GRU and an assistance-level instruction. Holding the cases and language model constant supported within-case comparisons of these context-and-policy configurations.

Risk-adaptive assistance in F4 used the 25th and 75th percentiles of validation-set calibrated risk. Cases below the lower threshold received a self-check-oriented instruction; middle-risk cases received a concept reminder and one minimal action; high-risk cases received a localized hint and the smallest supported next step. When the standard deviation across the five raw model probabilities exceeded 0.15, the instruction prioritized a verifiable check rather than a single causal diagnosis. This ensemble-disagreement rule served as an operational review heuristic.

All 544 outputs were generated with Qwen2.5-Coder-1.5B-Instruct \citep{hui2024qwen}. The same model, deterministic decoding procedure, and context limits were used throughout. Each response followed a compact structure containing an evidence-based observation, a feasible next action, a learner self-check, and an optional exact code excerpt. The instruction prohibited complete solutions and claims not supported by the supplied evidence.

Automated checks assessed structural completeness, actionability, self-check content, and whether quoted code matched the supplied source. Outputs with a structural error or unverifiable reference received one standardized repair attempt. Any source reference that remained unverifiable was removed, and unresolved structural cases were flagged for human review.

\subsection{Intervention Timing and Capacity Evaluation}

Two policy evaluations were conducted. The fixed-capacity analysis ranked the same labeled test pool by calibrated deep-model risk, risk from the best validation-selected model, consecutive-failure count, or attempt count. Budgets of 5\%, 10\%, 20\%, 30\%, 40\%, 50\%, 75\%, and 100\% of eligible states were evaluated. A random-alert baseline was estimated from 1000 draws at each budget. The principal policy measures were precision among selected states and the proportion of observed persistent failures captured. Ties were resolved using case identifiers without outcome information.

The sequential analysis represented a deployable trigger at each candidate state. A risk threshold was fixed at the 80th percentile of calibrated validation risk and applied unchanged to the test sequence. This policy was compared with alerting after every failure and after two consecutive failures. Evaluation included alert coverage, precision, persistent-failure capture, and observed time from the alert state to the next attempt. The 238 states with endpoint-censored primary outcomes still received predictions, while policy-performance denominators used only observed outcomes. The analysis therefore measures intervention prioritization across the recorded trajectories.

\subsection{Additional Analysis}

The source course included test-case feedback, natural-language feedback, and no-feedback conditions. Their associations with persistent failure were examined in the randomized course subset using a binomial generalized linear model with task and learner-state covariates and student-clustered standard errors. Exit-survey responses were summarized descriptively as learner-reported context. Analyses were conducted in Python 3.10.19 with PyTorch 2.5.1 \citep{paszke2019pytorch}.

\section{Results}

\subsection{Outcome Availability and Test-Set Composition}

The primary outcome was observed for 2755 of 2993 candidate states (92.1\%). In the held-out test partition, 555 states from 42 students had an observed primary outcome, including 222 (40.0\%) followed by persistent failure across the next two attempts. The remaining 42 test states retained predicted risk as endpoint-censored records outside supervised performance and policy denominators. An unmodified-retry outcome was available for 572 test states (180 positive; 31.5\%), near-term related-task performance for 209 states (97 successful; 46.4\%), and 7--28-day related-task performance for 67 states from 25 students (41 successful; 61.2\%).

\subsection{Prediction of Near-Term Persistent Failure}

Table~\ref{tab:primaryresults} reports primary test performance in validation PR-AUC order. Logistic regression was the validation-selected overall model (validation PR-AUC = 0.532). On held-out students, it achieved a PR-AUC of 0.550 (student-cluster 95\% CI [0.361, 0.676]), ROC-AUC of 0.681 [0.585, 0.741], and Brier score of 0.222. The MLP was the validation-selected deep model (validation PR-AUC = 0.525), with a test PR-AUC of 0.525 [0.352, 0.651], ROC-AUC of 0.670 [0.577, 0.732], and Brier score of 0.224. Overall, logistic regression provided the strongest validation-selected performance with competitive held-out discrimination and calibration.

\begin{table}[htbp]
\caption{Prediction of persistent failure on the student-disjoint test set. Confidence intervals were obtained by resampling students.\label{tab:primaryresults}}
\centering
\resizebox{\textwidth}{!}{%
\begin{tabular}{lccccc}
\toprule
Model & Validation PR-AUC & Test PR-AUC & PR-AUC 95\% CI & Test ROC-AUC & Brier \\
\midrule
Logistic regression & 0.532 & 0.550 & [0.361, 0.676] & 0.681 & 0.222 \\
MLP & 0.525 & 0.525 & [0.352, 0.651] & 0.670 & 0.224 \\
GRU fusion, multitask & 0.524 & 0.565 & [0.350, 0.699] & 0.689 & 0.230 \\
Transformer fusion & 0.519 & 0.547 & [0.345, 0.669] & 0.668 & 0.229 \\
GRU fusion & 0.518 & 0.551 & [0.350, 0.683] & 0.684 & 0.223 \\
Gradient boosting & 0.511 & 0.443 & [0.330, 0.547] & 0.575 & 0.251 \\
GRU, sequence only & 0.509 & 0.566 & [0.294, 0.686] & 0.628 & 0.234 \\
GRU fusion, no AST & 0.508 & 0.596 & [0.347, 0.715] & 0.697 & 0.227 \\
\bottomrule
\end{tabular}}
\end{table}

\begin{figure}[htbp]
\centering
\includegraphics[width=\textwidth]{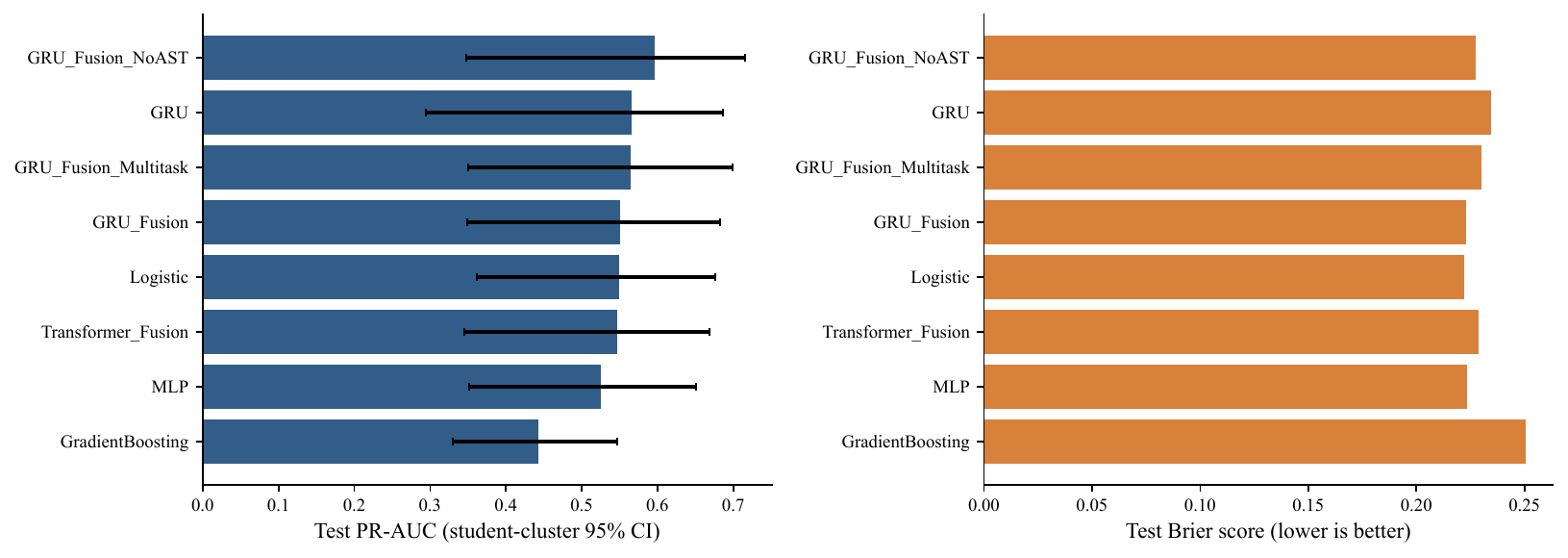}
\caption{Primary model performance on held-out students. Error bars show student-cluster 95\% confidence intervals. Models are ordered by validation PR-AUC; test results were not used for model selection.}
\label{fig:primarymodels}
\end{figure}

Several models achieved higher numerical test PR-AUC than their validation rankings suggested. The no-AST fusion GRU yielded the highest test PR-AUC (0.596) and ROC-AUC (0.697), despite a validation PR-AUC of 0.508. This finding is reported as an informative ablation result while retaining validation-based model selection. The multitask GRU achieved a test PR-AUC of 0.565 and ROC-AUC of 0.689, while the sequence-only GRU achieved PR-AUC 0.566 and ROC-AUC 0.628. Figure~\ref{fig:primarymodels} summarizes the estimates and student-cluster intervals across all eight configurations.

The selected MLP outperformed gradient boosting in the paired student bootstrap, with a PR-AUC difference of 0.082 and a 95\% interval of [0.009, 0.132]. Together with the logistic regression results, this comparison highlights the importance of benchmarking neural models against transparent baselines and matching model complexity to the predictive task.

\subsection{Feature Groups, Calibration, and Task Heterogeneity}

\begin{figure}[htbp]
\centering
\includegraphics[width=0.85\textwidth]{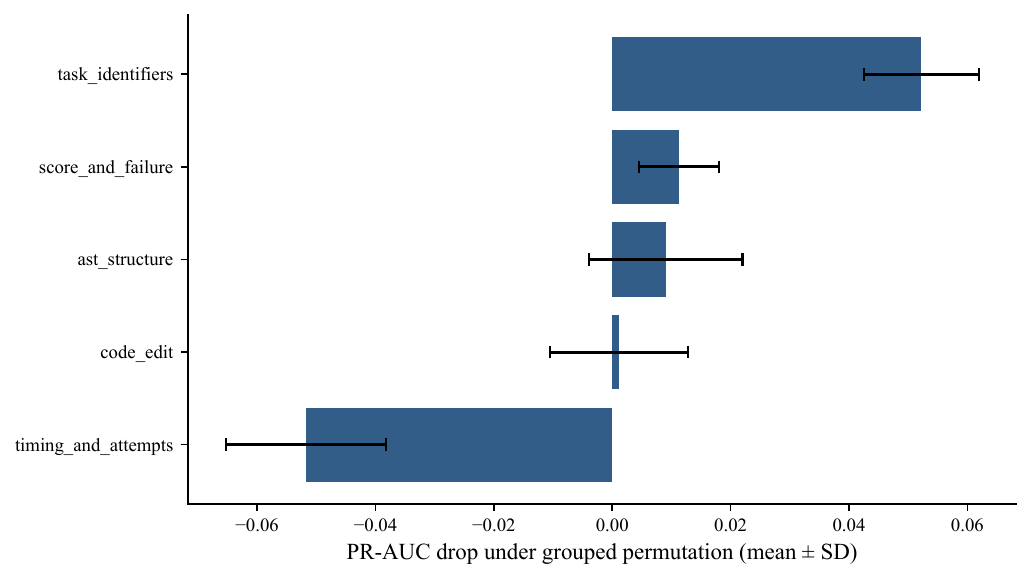}
\caption{Grouped permutation importance for gradient boosting. Values show changes in test PR-AUC across 30 permutations; negative values indicate improved test performance after permutation.}
\label{fig:importance}
\end{figure}

Grouped permutation analysis of the gradient-boosting model showed the largest mean PR-AUC decrease for task identifiers (mean decrease = 0.052, SD = 0.010). Score and failure features produced a decrease of 0.011 (SD = 0.007), followed by AST structure at 0.009 (SD = 0.013) and code-edit features at 0.001 (SD = 0.012). Timing and attempt features yielded a mean PR-AUC change of -0.052 (SD = 0.014), reflecting overlap and interaction with other feature groups. Figure~\ref{fig:importance} summarizes these group-level results.

The logistic model showed similar task dependence, with its largest absolute coefficients including specific function and laboratory indicators. Among continuous variables, consecutive failures had a positive standardized coefficient (0.429), whereas attempt count had a negative coefficient (-0.358); AST node count was positive (0.348), while AST return count was negative (-0.282). These conditional associations show how the model integrated task context with current learner behavior. Laboratory-level ROC-AUC ranged from 0.175 to 0.720 for the follow-up student-history model and from 0.429 to 0.714 for logistic regression among subgroups with at least 20 states. These results highlight meaningful heterogeneity across task contexts.

\subsection{Secondary Outcomes and Across-Task Student Histories}

Table~\ref{tab:secondaryresults} reports dedicated models trained separately for each secondary outcome. Unmodified retry was the most consistently predictable secondary behavior. Gradient boosting, which achieved the highest validation PR-AUC in the dedicated comparison, obtained a test PR-AUC of 0.599 and ROC-AUC of 0.735. With across-task history, the outcome-specific student GRU reached PR-AUC 0.631 and ROC-AUC 0.764, while the history-summary gradient-boosting model reached PR-AUC 0.659 and ROC-AUC 0.776. These results indicate that broader student history contributed useful information and that both sequential and summary representations captured this behavior.

\begin{table}[htbp]
\caption{Dedicated model performance for the three secondary outcomes.\label{tab:secondaryresults}}
\centering
\resizebox{\textwidth}{!}{%
\begin{tabular}{llrrrr}
\toprule
Outcome & Model & Test cases & Validation PR-AUC & Test PR-AUC & Test ROC-AUC \\
\midrule
Unmodified retry & Logistic regression & 572 & 0.395 & 0.502 & 0.654 \\
 & Gradient boosting & 572 & 0.445 & 0.599 & 0.735 \\
 & GRU fusion & 572 & 0.422 & 0.466 & 0.650 \\
\addlinespace
Near-term related-task performance & Logistic regression & 209 & 0.564 & 0.432 & 0.433 \\
 & Gradient boosting & 209 & 0.538 & 0.392 & 0.383 \\
 & GRU fusion & 209 & 0.608 & 0.424 & 0.303 \\
\addlinespace
Related-task performance at 7--28 days & Logistic regression & 67 & 0.679 & 0.825 & 0.691 \\
 & Gradient boosting & 67 & 0.650 & 0.820 & 0.699 \\
 & GRU fusion & 67 & 0.681 & 0.636 & 0.521 \\
\bottomrule
\end{tabular}}
\end{table}

Near-term related-task performance was the most challenging prediction target. The dedicated fusion GRU achieved validation PR-AUC 0.608 and test ROC-AUC 0.303; logistic regression reached ROC-AUC 0.433, and gradient boosting reached 0.383. The outcome-specific across-task GRU achieved validation PR-AUC 0.718, test PR-AUC 0.423, and ROC-AUC 0.302. These results highlight richer knowledge-component and instructional-context representations as an important direction for improving near-term transfer prediction.

The 7--28-day outcome showed promising longitudinal discrimination across 67 test cases from 25 students. Dedicated logistic regression and gradient boosting achieved ROC-AUC values of 0.691 and 0.699, respectively, with student-cluster intervals extending from approximately 0.49 to 0.85. The across-task multitask GRU reached PR-AUC 0.795 and ROC-AUC 0.683, with a ROC interval of [0.466, 0.857]. A history-summary logistic model achieved ROC-AUC 0.710 and PR-AUC 0.822. These estimates support further longitudinal validation with larger samples.

\begin{figure}[htbp]
\centering
\includegraphics[width=\textwidth]{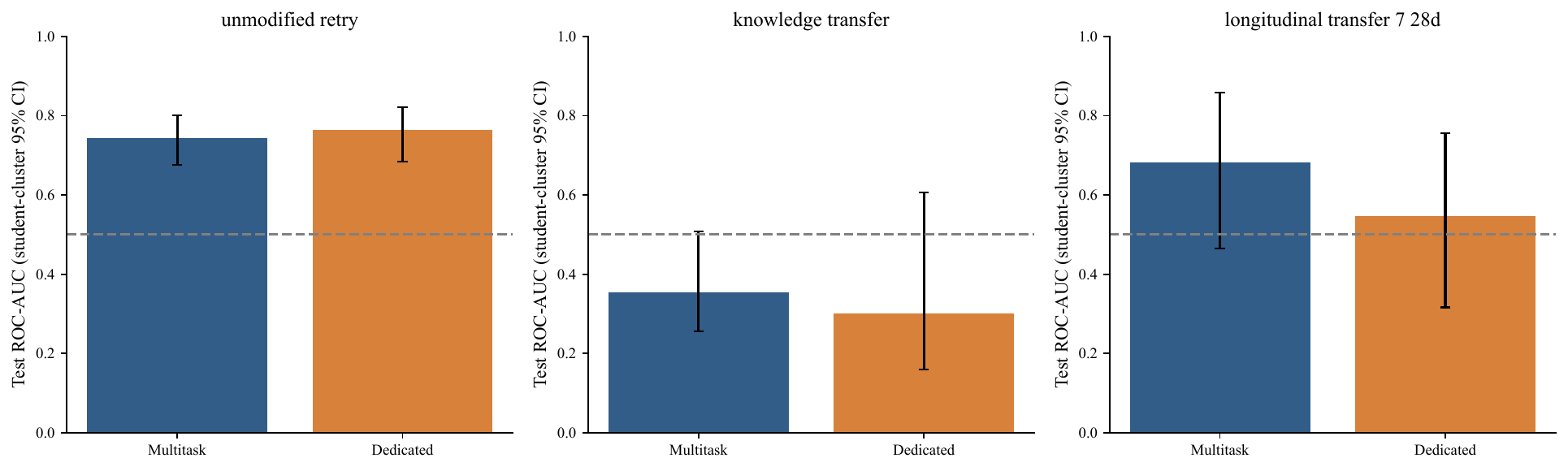}
\caption{Follow-up across-task student-history results. Bars show test ROC-AUC for the multitask and outcome-specific GRUs; error bars resample students.}
\label{fig:studenthistory}
\end{figure}

For the primary outcome, the follow-up across-task multitask GRU achieved validation PR-AUC 0.543, test PR-AUC 0.557, and ROC-AUC 0.669. Its PR-AUC differed from logistic regression by 0.008 in the paired student bootstrap, with a 95\% interval of [-0.101, 0.086]. Figure~\ref{fig:studenthistory} compares the shared multitask and outcome-specific models across the three secondary outcomes. The clearest improvement occurred for observable retry behavior, while the knowledge-related outcomes identify priorities for richer measurement.

\subsection{Recorded Feedback Scores and Newly Generated Feedback}

The recorded A--D feedback conditions each contained 136 human-reviewed scores. Mean scores increased monotonically from A (mean = 63.17, SD = 7.68) through B (71.17, SD = 7.68) and C (78.13, SD = 7.81) to D (86.13, SD = 7.81). Mean feedback lengths were 44.5, 71.1, 62.8, and 89.4 words, respectively. Paired mean differences were 8.0 points for B minus A, 15.0 for C minus A, 23.0 for D minus A, 15.0 for D minus B, and 8.0 for D minus C. These within-case contrasts summarize the recorded human assessments under the adopted rubric, including evidence and history dimensions.

The new generation procedure produced all 544 planned messages. Before repair, structurally valid responses were obtained for 96.3\% of F1 and F2 outputs, 91.9\% of F3 outputs, and 92.6\% of F4 outputs. All required components were present in 77.2\%, 89.0\%, 90.4\%, and 92.6\%, respectively. A self-check appeared in 96.3\% of F1 and F2 outputs, 91.9\% of F3, and 92.6\% of F4. Mean rendered length remained similar across conditions (59.5--60.9 words). No input exceeded the context limit, and one F4 output (0.7\%) reached the output limit.

\begin{figure}[htbp]
\centering
\includegraphics[width=\textwidth]{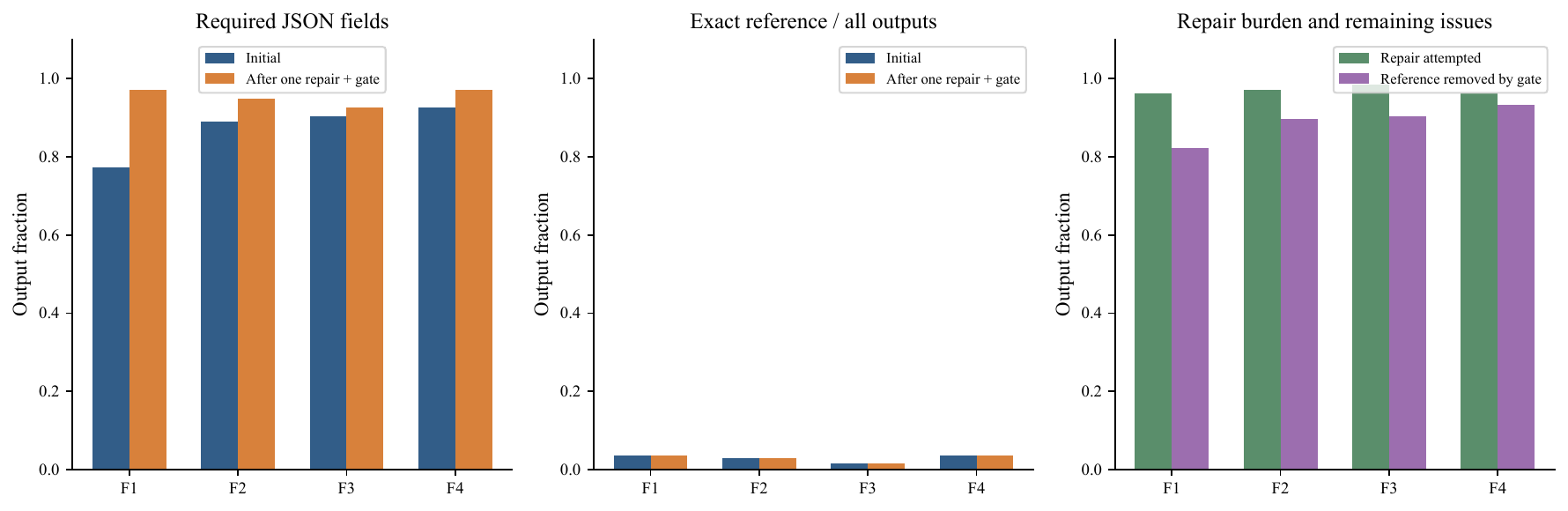}
\caption{Initial and final structural checks for newly generated feedback. Exact source-reference rates remained low, and the gate removed unverifiable references rather than retaining unsupported locations.}
\label{fig:feedbackrepair}
\end{figure}

The source-verification gate identified exact code matches in 5 F1, 4 F2, 2 F3, and 5 F4 messages, totaling 16 of 544 outputs (2.9\%). A standardized repair pass processed 528 outputs. After repair and deterministic reference gating, 519 outputs (95.4\%) contained all required components, while 25 required further structural review. Final condition-specific completeness rates were 97.1\% for F1, 94.9\% for F2, 92.6\% for F3, and 97.1\% for F4 (Table~\ref{tab:feedbackchecks}). The gate removed 484 unverifiable source references so that only exact matches were retained. Figure~\ref{fig:feedbackrepair} shows the initial and final checks.

\begin{table}[htbp]
\caption{Structural validation after one repair pass and deterministic reference gating. Rates are calculated within 136 cases per condition.\label{tab:feedbackchecks}}
\centering
\resizebox{\textwidth}{!}{%
\begin{tabular}{lrrrrrr}
\toprule
Condition & Initial complete & Repair attempted & Final complete & Exact reference & Reference removed & Manual review \\
\midrule
F1 & 77.2\% & 96.3\% & 97.1\% & 3.7\% & 82.4\% & 2.9\% \\
F2 & 89.0\% & 97.1\% & 94.9\% & 2.9\% & 89.7\% & 5.1\% \\
F3 & 90.4\% & 98.5\% & 92.6\% & 1.5\% & 90.4\% & 7.4\% \\
F4 & 92.6\% & 96.3\% & 97.1\% & 3.7\% & 93.4\% & 2.9\% \\
\bottomrule
\end{tabular}}
\end{table}

Risk-based instructions were assigned to all 136 F4 cases: 31 received the self-check level, 80 the concept-hint level, and 25 the localized-hint level. All cases remained below the prespecified ensemble-disagreement threshold. These results confirm that the prediction-to-assistance mapping operated as intended and yielded a complete set of messages for subsequent expert and learner evaluation.

\subsection{Intervention Capacity and Timing}

\begin{figure}[htbp]
\centering
\includegraphics[width=\textwidth]{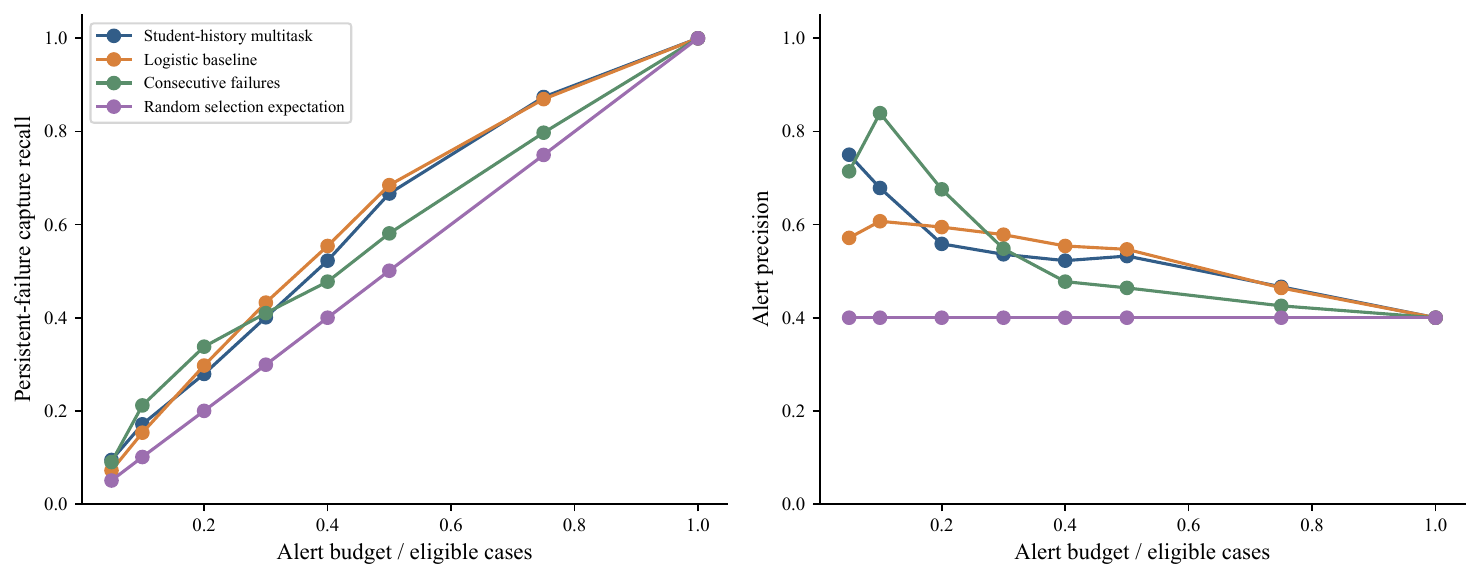}
\caption{Fixed-capacity comparison on the labeled test pool. Curves show precision and persistent-failure capture under recorded outcomes.}
\label{fig:policycapacity}
\end{figure}

Figure~\ref{fig:policycapacity} compares the across-task multitask policy with logistic risk, consecutive failures, and random selection under the same capacity. At a 5\% budget (28 alerts), the student-history policy captured 21 of 222 persistent failures, yielding precision 0.750 and capture recall 0.095. Logistic risk captured 16 cases (precision 0.571), while the consecutive-failure rule captured 20 (precision 0.714). At a 10\% budget, consecutive failures performed best, capturing 47 cases compared with 38 for the student-history model and 34 for logistic regression.

At the planned 20\% budget (111 alerts), the student-history policy captured 62 persistent failures (precision = 0.559; capture recall = 0.279). Logistic regression captured 66 (precision = 0.595; recall = 0.297), while consecutive failures captured 75 (precision = 0.676; recall = 0.338). The student-history minus logistic differences were -0.036 for precision (95\% student-bootstrap interval [-0.287, 0.116]) and -0.018 for capture recall [-0.149, 0.057]. At a 50\% budget, capture recall reached 0.667 for student history, 0.685 for logistic regression, and 0.581 for consecutive failures. These profiles show how policy performance varies with alert capacity and the desired balance between precision and capture.

For the sequential student-history policy, the validation-fixed threshold was 0.500. Applied to the test sequence, it triggered on 17.8\% of states, with precision 0.566 and capture recall 0.252. A sensitivity analysis using the validation-selected core deep model yielded a lower threshold of 0.492, 22.0\% coverage, precision 0.607, and capture recall 0.333. By comparison, alerting after every failure covered all states at the base precision of 0.400, while the two-consecutive-failures rule covered 97.7\% and captured 99.1\% of persistent failures. This rule reduced alert burden only marginally because most candidate states already occurred within repeated-failure sequences.

\begin{figure}[htbp]
\centering
\includegraphics[width=0.75\textwidth]{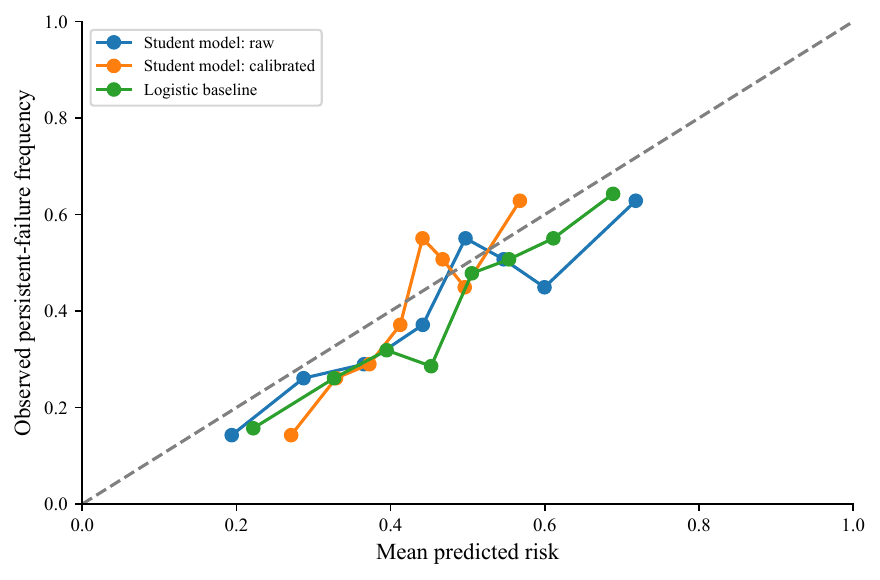}
\caption{Calibration of the across-task student-history model before and after validation-fitted scaling, with logistic regression for comparison.}
\label{fig:policycalibration}
\end{figure}

Validation-only calibration compressed the range of student-history probabilities and supported a fixed test-time threshold. Figure~\ref{fig:policycalibration} shows test reliability bins for the calibrated student-history model and logistic regression, providing a transparent basis for threshold selection.

\subsection{Historical Course Feedback and Survey Context}

\begin{figure}[htbp]
\centering
\includegraphics[width=0.9\textwidth]{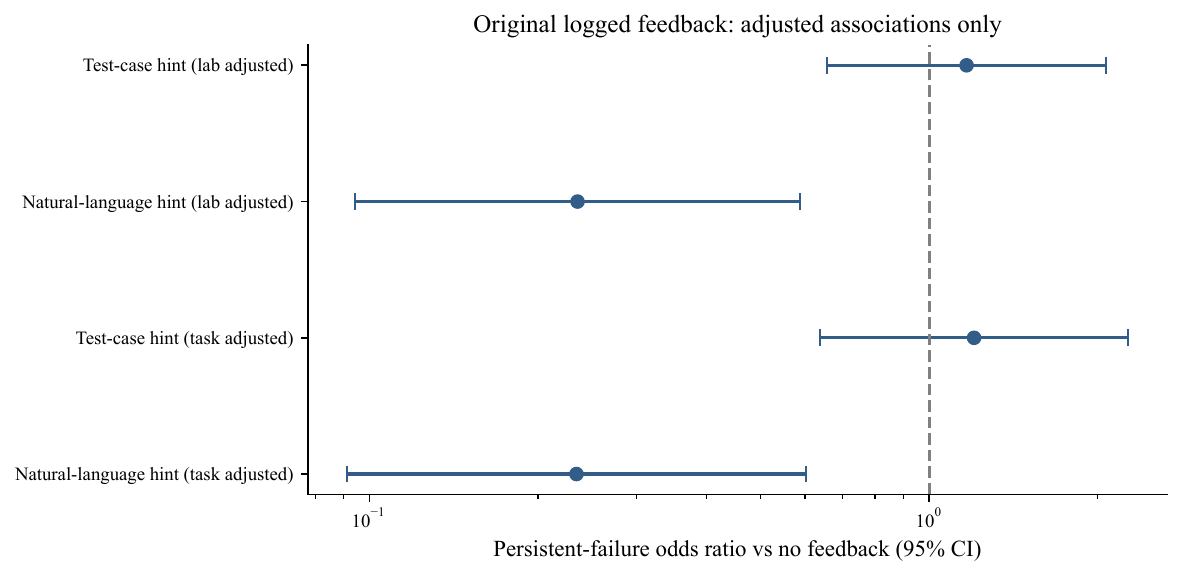}
\caption{Adjusted associations for recorded source-course feedback conditions. Intervals use student-cluster standard errors.}
\label{fig:historicalassociation}
\end{figure}

The historical analysis of the randomized course subset included 552 candidate states from 124 students. After adjustment for task, attempt count, consecutive failures, and edit fraction, the natural-language condition was associated with lower odds of subsequent persistent failure than the no-feedback condition in the task-fixed-effects model (odds ratio = 0.234, 95\% CI [0.091, 0.602]). The estimate was similar with laboratory fixed effects (odds ratio = 0.235 [0.094, 0.588]). The test-case condition yielded an odds ratio of 1.203 [0.639, 2.264] with task fixed effects and 1.166 [0.656, 2.073] with laboratory fixed effects (Figure~\ref{fig:historicalassociation}). These adjusted associations complement the predictive and generation analyses and inform evaluation of the newly generated F1--F4 messages.

Among 37 exit-survey respondents, 31 reported encountering generated feedback. Ratings were tabulated for all 37 respondents: 19 rated its helpfulness as 4 or 5 on the five-point scale, 13 selected 3, and five selected 1 or 2 (mean = 3.49). Fourteen rated learning attributed to feedback as 4 or 5, 12 selected 3, and 11 selected 1 or 2 (mean = 3.16). These ratings describe the full respondent sample.

Taken together, the results address the four research questions and demonstrate the feasibility of an auditable prediction--generation pipeline. Current and historical traces predicted persistent failure with moderate discrimination, with logistic regression providing a strong and interpretable benchmark. Broader histories improved prediction of observable resubmission behavior, while the 7--28-day analysis revealed a longitudinal performance signal. The generation pipeline produced concise, predominantly complete feedback structures after repair and routed unverifiable code references through explicit quality control. Risk-based policies concentrated persistent failures within restricted alert budgets, establishing an empirical basis for adapting assistance thresholds to instructional capacity.

\section{Discussion}

\subsection{Principal Findings}

This study integrated learning traces, predictive modeling, and evidence-constrained generation into an auditable system for personalized programming feedback. For RQ1, current-state and historical features predicted persistent failure with moderate discrimination under student-disjoint evaluation. The validation-selected logistic model achieved a test PR-AUC of 0.550 and ROC-AUC of 0.681, while unmodified retry and the 7--28-day outcome also showed useful predictive signal. Near-term related-task performance emerged as a distinct target requiring richer knowledge-component and instructional-context representations.

For RQ2, the comparison showed a practical balance between performance and parsimony. The validation-selected MLP performed similarly to logistic regression on the primary outcome, while recurrent models added value for observable resubmission behavior. Sequence models are therefore most useful when ordered history contributes stable predictive information, whereas logistic regression remains an efficient and transparent choice when discrimination and calibration are comparable.

For RQ3, standardized repair produced compact and structurally complete feedback in most cases. Source verification served as the central quality-control mechanism: unsupported code references were removed and unresolved outputs were routed for review. The matched four-condition design enables within-case comparison of contextual evidence and assistance instructions, providing a basis for expert and learner evaluation.

For RQ4, risk ranking concentrated persistent failures more effectively than random selection under restricted alert budgets. The sequential policy selected 17.8\% of eligible test states and captured 25.2\% of observed persistent failures, while a more sensitive configuration increased capture at higher coverage. These results support intervention thresholds that reflect available instructional capacity and the relative costs of missed and unnecessary support.

Together, the findings support a unified interpretation. Predictive models identify where support may be useful, calibrated policies determine when limited attention should be allocated, and evidence constraints govern what feedback may claim. Risk guides allocation, observable evidence grounds the message, and subsequent learner responses inform the next decision. The contribution therefore lies in coordinating prediction, decision, and generation around a defined learning process rather than optimizing prediction accuracy or message quality in isolation.

\subsection{Predictive Modeling, Measurement, and the Value of Learning Histories}

The competitiveness of logistic regression indicates that deep learning adds greatest value when an outcome benefits from temporal representation. Recurrent and attention-based models can capture evolving interaction histories \citep{abdelrahman2023knowledge}; here, broader histories were particularly informative for resubmission behavior, whereas current score, repeated failure, attempt history, and task identity carried substantial signal for the primary outcome. Comparing transparent and deep models therefore helps align model complexity with the educational target.

Task identifiers showed the largest positive grouped importance in the boosting model, indicating that failure patterns were strongly associated with programming context. This finding supports task-aware calibration and reinforces the use of concrete task, test, and code evidence in learner-facing feedback. Differences across secondary outcomes likewise favor outcome-specific representations. An unmodified retry can motivate a reflective prompt about monitoring and strategy change, whereas later related-task performance requires a broader account of learning opportunities. These distinctions follow recommendations to link behavioral traces to explicit constructs and temporal windows \citep{du2023traces}.

The later outcomes further clarify the value of learning history. Recent attempts directly characterize persistence and revision behavior, whereas performance on another task depends on content similarity, intervening instruction, practice opportunities, and time. A single learner representation is therefore unlikely to serve all outcomes equally well. Combining same-task sequences with broader histories provides a principled way to distinguish immediate debugging support from feedback intended to promote transfer or longer-term learning.

Student-disjoint partitions and decision-time prefixes reduced identity and temporal leakage \citep{kapoor2023leakage}, while validation-based selection preserved an independent final test. Calibration and capacity curves then translated predicted risk into alert volume, precision, and failure capture. This connection is important because intervention value depends on operational decisions as well as predictive discrimination.

\subsection{From Risk Scores to Evidence-Gated Progressive Assistance}

The findings support an \emph{evidence-gated progressive assistance} strategy built around four connected decisions. A timing gate combines calibrated risk with available support capacity. An evidence gate restricts feedback to observations available at that moment. An assistance gate adjusts specificity according to risk and recent behavior. A verification gate checks structure, source traceability, and model disagreement before delivery. Each subsequent attempt then provides new evidence for the next decision.

The F1--F4 conditions operationalize this strategy as a progressive sequence. At lower risk, feedback can prompt inspection of a test result or prediction of an output. Intermediate support adds a concept reminder and one minimal action. After repeated unsuccessful revisions, a localized hint can identify a relevant code region and suggest the smallest supported debugging step. These guardrails preserve independent reasoning while allowing assistance to become more specific when needed \citep{bastani2025guardrails}; the self-check component keeps learners responsible for verifying the explanation.

Evidence verification is essential because fluent messages can extend beyond the supplied record \citep{huang2025hallucination}. Code-specific claims therefore require a resolvable source anchor, while messages with conflicting evidence or unresolved structure enter focused review. This design directs instructor attention toward cases where professional judgment is most valuable.

Explainability should reflect its audience. Learners need a concise account of the observable issue, the relevance of the next action, and a way to verify progress. Instructors need access to the supporting trace, calibrated risk, uncertainty, and repair history. This layered design follows educational explainability research, which links explanation form to stakeholder and purpose \citep{khosravi2022explainable}. Timing can then preserve productive debugging by presenting recorded failures first and offering stronger support when help is requested or continuing difficulty is observed.

The timing policy also preserves learner agency. Immediate intervention is appropriate when repeated evidence indicates an unresolved impasse, while a short learner-controlled interval can support productive debugging. In practice, the system can first display recorded test evidence, then offer a scaffold on request or when the next attempt shows continuing difficulty. Capacity thresholds keep this sequence manageable by limiting intervention volume while preserving a clear escalation rule.

\subsection{Contributions to Learning Analytics and Generative Feedback Research}

The first contribution is a clear separation among prediction, decision, generation, and learning effects. Learning-analytics systems often present indicators without linking them to actionable strategies \citep{matcha2020dashboards,paulsen2024dashboards}. The present framework assigns distinct evidence to each claim: held-out outcomes assess prediction, policy replay assesses prioritization, and structural and provenance checks assess generated messages. This structure avoids treating plausible feedback as evidence of learning and can generalize to other settings that combine trace data with generative support.

The second contribution is the paired contextual design. The same 136 cases were represented under four context-and-policy configurations while the generator remained fixed. Prior studies established that language models can produce programming explanations, error messages, and hints at different levels \citep{sarsa2022generation,leinonen2023errors,xiao2024hints,lohr2025types}. This study extends that work by examining how learner history and calibrated risk enter generation and by producing matched messages suitable for blinded assessment.

The third contribution is capacity-aware intervention design. Fixed-capacity curves and sequential replay show how precision and failure capture change as support expands. Instructors can therefore align thresholds with available review time and the cost of missed support, linking learning analytics to feedback purpose and stakeholder action \citep{ifenthaler2020learning,banihashem2022analytics}. A staged implementation can begin with instructor review of risk, evidence, and escalation levels, followed by selective automation after local validation. This workflow preserves oversight while making personalized support operationally feasible.

The results also suggest a practical evaluation sequence. Institutions can first audit risk estimates and message evidence, then obtain blinded ratings of correctness, helpfulness, specificity, and solution disclosure before learner-facing deployment. Subsequent classroom evaluation can assess knowledge, self-regulated learning, transfer, and retention using measures aligned with each construct. This staged approach connects computational performance with educational outcomes without relying on a single metric to support every claim.

\section{Research Scope and Future Research}

The present study provides course-scale validation of the prediction--decision--generation pipeline through student-disjoint evaluation, time-valid features, paired feedback conditions, and explicit intervention budgets. Its operational outcomes and task-aware findings establish a basis for external validation across subsequent cohorts, institutions, programming languages, and assessment designs. Freezing feature definitions, calibration procedures, and decision thresholds before transfer will enable direct evaluation of model portability.

The next phase can extend this trace-based evaluation with learner-facing evidence. A prospective comparison of usual feedback, fixed evidence-based feedback, and evidence-gated progressive assistance can combine an immediate parallel-item test, a structurally different transfer task, an unassisted delayed test after two to four weeks, and a validated self-regulated-learning measure. Blinded programming-education raters can assess correctness, grounding, actionability, clarity, and solution disclosure, while subgroup calibration, accessibility, alert exposure, and learner autonomy can inform equitable implementation.

A student-level or micro-randomized design can examine repeated intervention opportunities using multilevel models. Preregistering prediction models, thresholds, prompts, outcomes, and moderation analyses will connect the computational framework to direct measures of knowledge, transfer, self-regulation, and retention. This research agenda can establish how the strategy generalizes and which combinations of timing, evidence, and assistance most effectively support learning.

\section{Conclusions}

This study developed and evaluated a framework integrating learning traces, calibrated risk prediction, constrained feedback generation, and intervention timing. Using 2993 failed programming states from 215 students, the framework predicted persistent difficulty, generated four matched forms of contextualized feedback, and evaluated alert policies under limited intervention capacity. Its central principle is that prediction, decision, generation, and learning outcomes require distinct forms of evidence.

The results showed that current performance and prior behavior provided meaningful predictive value. Logistic regression offered a strong interpretable benchmark, while deep and broader-history models contributed outcome-specific information, particularly for observable resubmission behavior. The generation pipeline produced predominantly complete feedback structures after repair, and exact source gating provided a transparent quality-control mechanism. Risk-based policies concentrated persistent-failure cases within restricted alert budgets and supported threshold selection according to instructional capacity.

Based on these findings, the study proposes evidence-gated progressive assistance: calibrated risk informs whether support is triggered, time-valid evidence constrains message content, assistance progresses from self-checks to localized guidance, and selected outputs receive focused review. This strategy provides a practical pathway from learning analytics to personalized, explainable, and timely feedback by aligning model complexity and feedback specificity with observed learner needs.

The study establishes a technical and methodological foundation for prospective educational evaluation. Future work can assess knowledge, self-regulated learning, transfer, and long-term retention through learner deployment, validated measures, blinded feedback assessment, and delayed testing. This framework positions generative AI and learning analytics as an auditable support process that targets assistance to relevant moments while preserving learner reasoning and instructor oversight.

\authorcontributions{Conceptualization, S.W.; methodology, S.W.; software, S.W.; validation, S.W.; formal analysis, S.W.; investigation, S.W.; data curation, S.W.; writing---original draft preparation, S.W.; writing---review and editing, S.W.; visualization, S.W.; project administration, S.W. The author has read and agreed to the published version of the manuscript.}

\funding{This research received no external funding.}

\institutionalreview{Ethical review was not required for this secondary analysis of publicly available, de-identified data.}

\informedconsent{The source dataset documents participants' consent for research use; no new participants were recruited.}

\dataavailability{The source data are publicly available through the ProgFeed Dataset repository cited in this article \citep{umass2025progfeed}. The analytical records and results supporting the findings are described in the article. Further inquiries can be directed to the corresponding author.}

\conflictsofinterest{The author declares no known competing financial interests or personal relationships that could have appeared to influence the work reported in this paper.}

\begin{adjustwidth}{-\extralength}{0cm}
\reftitle{References}
\bibliography{references}
\PublishersNote{}
\end{adjustwidth}

\end{document}